\documentclass[conference]{IEEEtran}
\IEEEoverridecommandlockouts

\usepackage{cite}
\usepackage{amsmath,amssymb,amsfonts}
\usepackage{algorithmic}
\usepackage{graphicx}
\usepackage{textcomp}
\usepackage{xcolor}
\usepackage{booktabs}
\usepackage{url}
\def\BibTeX{{\rm B\kern-.05em{\sc i\kern-.025em b}\kern-.08em
    T\kern-.1667em\lower.7ex\hbox{E}\kern-.125emX}}
\begin{document}

\title{Enabling Streaming User Transcription in Full-Duplex Speech-to-Speech Models}


\author{
\IEEEauthorblockN{
Ke Hu, Nourchene Ferchichi, Edresson Casanova, Ankita Pasad,\\
Elena Rastorgueva, Chen Chen, Nithin Rao Koluguri, Piotr Zelasko,\\
Yifan Peng, Hainan Xu, Zhehuai Chen, Boris Ginsburg
}
\IEEEauthorblockA{
NVIDIA\\
kevinhu@nvidia.com
}
}

\maketitle

\begin{abstract}
    Full-duplex speech-to-speech (S2S) models enable natural conversational AI by allowing simultaneous listening and speaking. However, these models typically lack inherent user speech transcription, which is essential for applications such as conversation logging, accessibility features, and quality monitoring. In this work, we propose an efficient method to add streaming ASR capabilities to an existing duplex S2S model by introducing a lightweight ASR head in parallel to the agent text head. Our approach requires minimal additional parameters and no significant architectural changes to the base S2S model, enabling real-time user transcription while preserving full-duplex conversational capabilities including turn-taking and barge-in handling. Experimental results demonstrate that our method achieves streaming average WER of 10.21\% on the HuggingFace Open ASR Leaderboard within the duplex S2S framework. Additionally, we show that the same architecture trained as a standalone streaming ASR model achieves competitive results (7.73\% WER) compared to current SOTA models. We will open-source our training and inference code to facilitate further research in joint streaming ASR and S2S modeling.
\end{abstract}

\section{Introduction}

Streaming automatic speech recognition (ASR) is fundamental to real-time human-computer interaction, enabling applications such as live captioning, voice assistants, and conversational AI. As large language models (LLMs) \cite{brown2020language, dubey2024llama, openai2024gpt4o, qwen2025qwen3, anthropic2024claude3, deepseek2025v32, guo2025deepseek, minimax2025m01, muralidharan2024minitron, nvidia2025nemotron_nano2} have transformed natural language processing, there is growing interest in extending these capabilities to speech or multimodal inputs. Recent work has explored adapting LLMs to process speech or multimodal inputs for various tasks \cite{chen2024salm, openai2024gpt4o, team2025gemini25, qwen2025omni, qwen2026asr, wang2024speechllm, wang2024freeze, zeng2024glm}. A fully capable conversational speech system benefits from not only agent response generation but also streaming user transcription capabilities alongside features such as turn-taking and barge-in handling.

There have been a number of works on incorporating speech inputs to LLMs. Traditional spoken dialogue systems cascade ASR, LLM, and TTS modules \cite{huang2024audiogpt}, and while they naturally provide user transcriptions, this approach has potentially higher latency and makes it difficult to incorporate paralinguistic information in S2S modeling. This has motivated research into end-to-end speech-to-speech (S2S) models. Initial efforts focused on half-duplex, turn-based interactions \cite{zhang2023speechgpt, kim2024unified, zeng2024glm, qwen2025omni, wang2024freeze, xie2024mini, intrinsicvoice2024, llamaomni2024}, which still fail to capture the interactive nature of real dialogue. Some systems like gpt-realtime \cite{openai2024gpt4o, openai2025realtime}, Freeze-Omni \cite{wang2024freeze}, FireRedChat \cite{fireredchat2024}, and FlexDuo \cite{flexduo2025} achieve low-latency interactions through external voice activity detection and turn-taking modules, but fundamentally rely on explicit turn detection rather than simultaneous processing. Recently, full-duplex S2S models have emerged that enable simultaneous listening and speaking. Moshi \cite{defossez2024moshi} and PersonaPlex \cite{roy2026personaplex} are full-duplex conversational models that jointly model both user and agent audio streams with depth-based attention mechanisms, but require extensive speech-text pretraining from scratch and do not provide streaming user transcription capabilities. Other approaches like SyncLLM \cite{yusalmonn} and OmniFlatten \cite{zhang2024omniflatten} also achieve full-duplex conversation through various architectural designs but share the same limitation of lacking explicit user transcription outputs. Recent work has explored augmenting duplex speech LLMs with chain-of-thought reasoning by interleaving ASR and reasoning tokens within a single text monologue stream \cite{shih2025speech_llm_think}, though this conflates transcription and reasoning into a shared channel rather than treating ASR as a dedicated output.

A recent duplex S2S architecture \cite{salm_duplex, asru_duplex} demonstrated that any text LLM can be converted into a full-duplex conversational agent without requiring extensive speech-text pretraining, by using a pretrained streaming encoder for user input and parallel text and audio heads for agent output. However, this architecture does not provide explicit user speech transcription, which is valuable for downstream applications such as conversation logging, accessibility features, and quality monitoring.

In this work, we build on this duplex S2S architecture by introducing a streaming ASR head in parallel to the agent text head, enabling the model to perform continuous speech recognition while maintaining full-duplex conversational capabilities. Our approach enables frame-level streaming ASR within a decoder-only LLM architecture, allowing simultaneous speech recognition and agent response generation. We further demonstrate that the same architecture can be trained as a standalone streaming ASR model, achieving competitive results on standard benchmarks.

Our main contributions are as follows:
\begin{itemize}
    \item We propose an efficient method to add streaming ASR capabilities to a full-duplex S2S model, requiring minimal additional parameters while preserving turn-taking and barge-in performance.
    \item We demonstrate that the S2S model achieves streaming ASR capability, enabling real-time user transcription alongside agent response generation.
    \item We show that the same architecture trained as a standalone ASR model achieves competitive results on the HuggingFace Open ASR Leaderboard~\cite{hf_open_asr_leaderboard}.
    \item We will open-source our training and inference code to facilitate reproducibility and enable further research in joint streaming ASR and S2S modeling.
\end{itemize}

\section{Related work}

Recent advances in streaming ASR have led to several notable systems. Encoder-based models such as FastConformer \cite{nvidia2023stt_fastconformer, nvidia2023stt_fastconformer_multi} and Parakeet \cite{parakeet} achieve state of the art performance using conformer blocks with cache-based inference, but it is unclear how to incorporate them into a full-duplex S2S model. Kyutai STT \cite{kyutai2024stt} achieves strong streaming ASR performance but lacks the conversational capabilities required for a full-duplex framework. LLM-based approaches such as Qwen-ASR \cite{qwenasr} leverage the language understanding of large models for speech recognition, but rely on chunk-based processing where minimum latency is determined by chunk size, making integration with continuously streaming S2S models prohibitively complex.

Separately, significant progress has been made in full-duplex spoken dialogue modeling. Moshi \cite{defossez2024moshi} and PersonaPlex \cite{roy2026personaplex} enable simultaneous listening and speaking by jointly modeling user and agent audio streams, but require extensive speech-text pretraining from scratch and do not provide streaming user transcription. Systems like Freeze-Omni \cite{wang2024freeze}, FireRedChat \cite{fireredchat2024}, and FlexDuo \cite{flexduo2025} reduce interaction latency through external voice activity detection and turn-taking modules, but fundamentally rely on explicit turn detection rather than truly simultaneous processing. SALM-Duplex \cite{salm_duplex} and its system demonstration \cite{asru_duplex} convert any pretrained text LLM into a full-duplex agent without requiring speech pretraining, but similarly lack explicit user transcription output. A related line of work explores augmenting duplex speech LLMs with transcription by interleaving ASR and reasoning tokens within a single text stream \cite{shih2025speech_llm_think}, but this conflates transcription and reasoning into a shared channel rather than treating ASR as a dedicated parallel output. Our work addresses this by adding a dedicated streaming ASR head to the SALM-Duplex architecture, enabling real-time user transcription alongside full-duplex conversation capabilities.

\section{Model Architecture}

\begin{figure}[t]
\centering
\includegraphics[width=\linewidth]{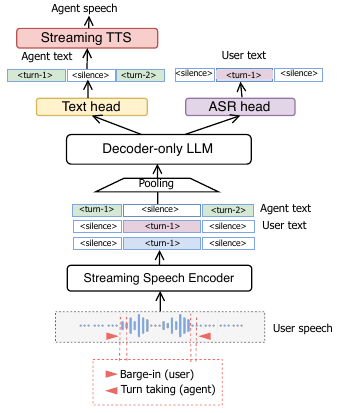}
\caption{Our architecture for adding a streaming ASR head to the speech-to-text part of the S2S duplex model. The model takes continuous user audio embeddings, previous ASR and agent text tokens as inputs, and outputs user ASR and agent text in parallel.}
\label{fig:architecture}
\end{figure}

\subsection{Full-Duplex S2S Model}

Our work builds upon the duplex speech-to-speech (S2S) architecture proposed in \cite{salm_duplex, asru_duplex} by adding a streaming ASR head. As shown in Figure~\ref{fig:architecture}, our extended model processes three input streams: user speech, user transcript, and agent text. User speech is first encoded by a 600M parameter Parakeet streaming speech encoder \cite{parakeet}, generating continuous embeddings at an 80ms frame rate. The backbone LLM is initialized from NVIDIA Nemotron-Nano-9B-v2-Base \cite{nvidia2025nemotron}, a 9 billion parameter decoder-only language model optimized for reasoning and instruction-following tasks. The generated user and agent text tokens are autoregressively fed back as inputs to the backbone LLM.

The model is trained with multi-channel next token prediction, simultaneously generating user text and agent text through parallel heads. The user and agent embeddings are time-aligned and added before being passed to the decoder-only LLM, together with input user speech encoding. We use equal loss for user and agent text prediction. Unlike streaming ASR, agent text is predicted without word-level alignment to give the user a preview of the agent response before speech generation finishes. Our agent text with turn taking information is then fed to a streaming TTS \cite{casanova2026voicechat} to generate agent speech. In this work, to incorporate user speech transcription, we focus on modifying the speech-to-text part of the architecture.

In this work, our training includes a pretraining stage followed by supervised fine-tuning (SFT). During pretraining, the model is trained on interleaved speech-to-text conversation data. In the SFT stage, we fine-tune the model on a mixture of diverse data sources including multi-turn conversational data and ASR transcription data. To improve robustness to diverse acoustic conditions, we apply background noise augmentation during the SFT stage with 0.5 probability, where additive noise is randomly selected from a collection of over 60,000 noise files. The signal-to-noise ratio (SNR) is randomly sampled in a wide range to enable the model to handle various acoustic environments.

\subsection{Streaming ASR Head}

We augment the duplex S2S architecture by introducing a streaming ASR module in parallel to the existing agent text head. This module consists of both a separate embedding layer and a prediction head for user transcription. Both layers are initialized from the original corresponding LLM backbone layers. As illustrated in Figure~\ref{fig:architecture}, the ASR head takes the LLM's hidden states and predicts user transcription tokens in a streaming fashion, while the separate embedding layer allows the model to learn user text representations independently from the agent text embeddings. This design enables real-time speech recognition concurrent with agent response generation.

The streaming ASR head shares the same LLM backbone as the agent text head, allowing it to leverage the contextual understanding from the conversation flow. Only a single decoding pass is needed to jointly produce both user and agent texts. During training, we add an ASR loss term using next token prediction that supervises the user transcription output alongside the agent text loss. This joint training enables the model to perform streaming ASR while maintaining its full-duplex conversational capabilities. For the standalone streaming ASR configuration, we train the model without the agent text heads, focusing solely on the streaming speech recognition task.

\subsection{On-the-fly (OTF) Forced Alignment}

To enable streaming ASR training, we require frame-level alignment between user speech and transcription text. We use the torchaudio CTC-based forced alignment API with the MMS-FA acoustic model \cite{pratap2024scaling} for on-the-fly (OTF) forced alignment during training. Given user speech audio and transcripts, the forced aligner produces word-level timestamps that are used to align text with speech. We then use the timestamps to align text tokens with speech frames at the start of each word.

To improve streaming ASR quality, we introduce a user text delay $d_u$ that shifts the transcription targets forward in time relative to the speech frames. On the other hand, to facilitate agent turn taking, we apply a separate agent text delay $d_a$ to help the agent learn reliable timing to respond. The delays $d_u$ and $d_a$ are hyperparameters that control the trade-off between streaming ASR latency and turn taking accuracy.

For word-level alignment, we experimented with both left alignment (text tokens aligned to the start of each word) and right alignment (text tokens aligned to the end of each word). We found that left alignment yields better performance, presumably because speech onset is easier to detect in this setup. Between consecutive words, we use pad tokens to fill the frames where no text prediction is required. For example, the phrase ``hello world'' with left-aligned tokens would produce the target sequence like: ``\_hel lo \texttt{<pad>} \texttt{<pad>} \_world \texttt{<pad>} \texttt{<pad>}'', where the underscore denotes word boundaries and \texttt{<pad>} tokens fill the remaining frames within each word's duration. We also experimented with using a distinct end-of-word token instead of the pad token at word boundaries, but did not observe a significant difference in performance.

\section{Experiments}

\subsection{Data}

Our training proceeds in two stages: pretraining and supervised fine-tuning (SFT). In the pretraining stage, the model is trained on interleaved speech-to-text data to learn the fundamental knowledge of user and agent conversations. In the SFT stage, we train on a mixture of diverse data sources including: interleaved S2S data, text-to-text conversations, multi-turn conversational SFT data, multiple-choice question answering, single-turn speech instruction data, and ASR training data. Background noise augmentation is applied during the SFT stage only. The multi-turn conversational SFT data is synthesized following the approach described in \cite{salm_duplex}. The text-to-text data helps maintain the language modeling capabilities of the backbone LLM. The ASR data provides additional supervision for the streaming ASR head and consists of two types of data: Open-source and publicly available ASR training data including LibriSpeech \cite{panayotov2015librispeech}, VoxPopuli \cite{wang2021voxpopuli}, Common Voice \cite{ardila2020common}, VCTK \cite{yamagishi2019vctk}, SPGISpeech \cite{oneill2021spgispeech}, etc, as well as our in-house training data. We use the English portions of the ASR training data, totaling 16k hours. The signal-to-noise ratio (SNR) is randomly sampled between -30 dB and 60 dB to enable the model to handle various acoustic environments.

For the standalone streaming ASR experiments, we additionally leverage English data from Granary \cite{koluguri2025granary}, which combines open-source Creative Commons speech corpora including YODAS (YouTube-Oriented Dataset for Audio and Speech) and YouTube-Commons (YTC). The dataset enhances quality through a pseudo-labeling pipeline with segmentation, two-pass ASR inference, and hallucination filtering.

We evaluate our models on two types of test sets. For streaming ASR evaluation, we use benchmarks from the HuggingFace Open ASR Leaderboard \cite{hf_open_asr_leaderboard}, including \texttt{LibriSpeech} \texttt{test-clean} and \texttt{test-other}, \texttt{SPGISpeech}, \texttt{GigaSpeech}, \texttt{Earnings22}, \texttt{AMI}, \texttt{TED-LIUM}, and \texttt{VoxPopuli}. For turn-taking and conversational evaluation, we use Full-Duplex-Bench V1 (FDB-v1) \cite{lin2025fullduplexbench} and an internal test set of interactions with the model in real-world setups and containing around 60 multi-turn conversations covering diverse topics, with each conversation containing roughly 4 turns. The recordings were made in various acoustic environments using different devices and headsets to ensure robustness evaluation. To create the dataset, conversation scripts were first generated using a text LLM, then users recorded their turns while simulating natural conversation flow by allowing pauses for agent responses. Multiple conversations were recorded per topic to ensure diversity.

\subsection{Turn Taking and Streaming ASR}

\begin{table*}[!t]
\caption{Streaming ASR results (WER \%, $\downarrow$) in S2S model.}
\label{tab:duplex_asr}
\centering
\setlength{\tabcolsep}{10pt}
\renewcommand{\arraystretch}{1.3}
\begin{tabular}{l ccccccccc}
\toprule
Model & LS-clean & LS-other & SPGI & Giga & Earn22 & AMI & Tedlium & Voxpop & Avg \\
\midrule
Ours & 3.9 & 8.48 & 4.95 & 14.22 & 16.87 & 18.36 & 5.98 & 8.9 & 10.21 \\
\bottomrule
\end{tabular}
\vspace{-1em}
\end{table*}

\begin{table}[!t]
\caption{Turn-taking and barge-in evaluation. Precision and recall are in \%, and latencies are in ms.}
\label{tab:duplex_tt}
\centering
\begin{tabular}{l ccccc}
\toprule
& \multicolumn{3}{c}{Turn-taking} & \multicolumn{2}{c}{Barge-in} \\
\cmidrule(lr){2-4} \cmidrule(lr){5-6}
Model & Pr $\uparrow$ & Rec $\uparrow$ & Lat $\downarrow$ & Acc $\uparrow$ & Lat $\downarrow$ \\
\midrule
Baseline (noASR) & 86.1 & 96.9 & 410 & 100 & 393 \\
Ours & 90 & 95 & 431 & 100 & 374 \\
\bottomrule
\end{tabular}
\vspace{-1em}
\end{table}

\begin{table}[!t]
\caption{Intelligence evaluation.}
\label{tab:duplex_intel}
\centering
\begin{tabular}{l ccc}
\toprule
Model & OpenbookQA (\%) $\uparrow$ & AE (/5) $\uparrow$ & CE (/5) $\uparrow$ \\
\midrule
Moshi \cite{defossez2024moshi, voicebench} & 26.15 & 2.01 & 1.60 \\
Qwen2-Audio \cite{voicebench} & 67.91 & 4.11 & 3.77 \\
\midrule
Baseline (noASR) & 66.59 & 3.71 & 3.24 \\
Ours & 69.01 & 3.83 & 3.11 \\
\bottomrule
\end{tabular}
\vspace{-1em}
\end{table}

\begin{table*}[!t]
\caption{FDB-v1 \cite{lin2025fullduplexbench} evaluation. TORs are in \%, latencies are in ms, and the GPT score is out of 5.}
\label{tab:fdb}
\centering
\setlength{\tabcolsep}{8pt}
\renewcommand{\arraystretch}{1.0}
\begin{tabular}{l cc ccc c}
\toprule
& \multicolumn{2}{c}{Smooth TT} & \multicolumn{3}{c}{User Interruption} & Pause \\
\cmidrule(lr){2-3} \cmidrule(lr){4-6} \cmidrule(lr){7-7}
Model & TOR $\uparrow$ & Lat $\downarrow$ & TOR $\uparrow$ & GPT $\uparrow$ & Lat & TOR $\downarrow$ \\
\midrule
Moshi \cite{defossez2024moshi, lin2025fullduplexbench} & 94 & 265 & 100 & 0.77 & 257 & 98 \\
Baseline (noASR) & 95.15 & 257 & 92.5 & 4.38 & 369 & 51.4 \\
Ours & 96.12 & 477 & 94 & 3.99 & 355 & 44.4 \\
\bottomrule
\end{tabular}
\end{table*}

\begin{table*}[!t]
\caption{Standalone streaming ASR results (WER \%, $\downarrow$). Numbers in parentheses indicate latency in training.}
\label{tab:asr_results}
\centering
\renewcommand{\arraystretch}{1.0}
\begin{tabular}{l cc cccccc c}
\toprule
Model & LS-clean & LS-other & SPGI & Giga & Earn22 & AMI & Tedlium & Voxpop & Avg \\
\midrule
FastConformer-80ms \cite{nvidia2023stt_fastconformer} & 2.57 & 6.31 & 6.16 & 14.92 & 21.03 & 28.37 & 6.17 & 8.12 & 11.71 \\
FastConformer-multi (1.12s) \cite{nvidia2023stt_fastconformer_multi} & 2.19 & 5.32 & 5.76 & 14.47 & 21.45 & 27.85 & 5.70 & 7.42 & 11.27 \\
Nemotron-Speech-0.6B (1.12s) \cite{nvidia2025nemotron_speech_streaming} & 2.31 & 4.75 & 2.62 & 11.45 & 12.48 & 11.58 & 4.50 & 7.57 & 7.16 \\
Qwen3-ASR-1.7B (2s) \cite{qwen2026asr,hf_open_asr_leaderboard} & 1.63 & 3.4 & 2.84 & 8.74 & 10.25 & 10.56 & 2.28 & 6.35 & 5.76 \\
Qwen3-ASR-0.6B (2s) \cite{qwen2026asr,hf_open_asr_leaderboard} & 2.13 & 4.45 & 3.03 & 9.14 & 11.06 & 11.66 & 2.85 & 7.07 & 6.42 \\
Kyutai STT-2.6B (2.5s) \cite{kyutai2024stt,hf_open_asr_leaderboard} & 1.70 & 4.32 & 2.03 & 9.81 & 10.99 & 12.17 & 3.35 & 6.79 & 6.40 \\
\midrule
Ours (1.6s) & 2.68 & 6.04 & 4.87 & 11.64 & 15.01 & 14.20 & 4.61 & 8.70 & 8.47 \\
\quad + YODAS and YTC \cite{koluguri2025granary} & 2.48 & 6.03 & 3.66 & 11.21 & 13.56 & 12.97 & 4.03 & 7.91 & 7.73 \\
\bottomrule
\end{tabular}
\end{table*}

We evaluate the streaming ASR performance of our model when integrated into the full-duplex S2S framework. For all experiments, we use a user text delay of $d_u = 1.2$s and an agent text delay of $d_a = 0.16$s. The choice of these hyperparameters is to achieve a balance between reasonable ASR performance and immediate agent response. Tables~\ref{tab:duplex_asr}, \ref{tab:duplex_tt}, and \ref{tab:fdb} present the complete evaluation results for our duplex S2S model with the integrated streaming ASR head, including streaming ASR performance on the HuggingFace Open ASR Leaderboard \cite{hf_open_asr_leaderboard}, turn-taking metrics, intelligence scores, and FDB-v1 \cite{lin2025fullduplexbench} results.

As shown in Table \ref{tab:duplex_asr}, our model achieves 10.21\% average WER while simultaneously supporting agent response generation and full-duplex conversation. This is better than the \texttt{FastConformer-80ms} (11.71\%) and \texttt{FastConformer-multi} (11.27\%) models (Table~\ref{tab:asr_results}), which are dedicated streaming ASR models, demonstrating that our approach achieves competitive ASR performance even within the duplex S2S framework. 

For turn-taking, we report precision (Pr), recall (Rec), and latency (Lat.) for regular turn-taking, as well as barge-in accuracy (Acc) and barge-in latency (Lat.) in the Barge-in column using our internal test set. For intelligence, we report OpenbookQA accuracy, AlpacaEval (AE), and CommonEval (CE) scores from VoiceBench \cite{voicebench}. The turn-taking metrics are computed by extracting user speech segments using voice activity detection (VAD) \cite{silero_vad}, while agent response segments are derived from the model's predicted text with \texttt{<bos>} and \texttt{<eos>} timestamps. \textbf{Precision} measures the proportion of agent turns correctly following user turns, where an agent turn is a true positive if it starts within 1s before to 1.5s after a user segment ends. \textbf{Recall} measures the proportion of user utterances that receive an agent response starting within 1.5s. These thresholds are chosen empirically to align with our subjective experience but one can adjust them and we also compute turn taking latency as a complementary metric. \textbf{Latency} is the average time delay between user speech ending and agent response starting for correctly matched turns. For barge-in evaluation, a barge-in event is detected when the user starts speaking while the agent is still speaking. \textbf{Barge-in accuracy} is the percentage of barge-in events where the agent successfully stops speaking within 1.5s after the user interruption. \textbf{Barge-in latency} is the average time for the agent to stop after a successful barge-in. 

As shown in Table \ref{tab:duplex_tt}, we have achieved competitive turn taking performance (90\% precision and 95\% recall, with 431ms latency) based on our internal test set, and 100\% barge-in accuracy with 374ms barge-in latency. As shown in Table~\ref{tab:duplex_intel}, our model also achieves an AlpacaEval (AE) score of 3.83 and a CommonEval (CE) score of 3.11 (out of 5), with OpenbookQA accuracy of 69.01\%. This represents a substantial improvement over Moshi \cite{defossez2024moshi}. Compared to Qwen2-Audio \cite{voicebench}, a turn-based audio LLM, our model achieves slightly higher OpenbookQA accuracy (69.01\% vs.\ 67.91\%) but shows a gap in CommonEval (3.11 vs.\ 3.77).

Table~\ref{tab:fdb} shows the FDB-v1 \cite{lin2025fullduplexbench} results comparing our model against Moshi \cite{defossez2024moshi}. FDB-v1 evaluates three key interactive behaviors: smooth turn-taking, user interruption handling, and pause handling (we use the Candor set). Our model achieves better smooth turn-taking (TOR 96.12\% vs. 94\%) and worse user interruption TOR (94\% vs. 100\%) to Moshi, but a notably higher GPT score (3.99 vs. 0.77), indicating significantly better response quality upon interruption. For pause handling, our model produces fewer false takeovers (TOR 44.4\% vs. 98\%). The higher smooth turn-taking latency of our model (477ms vs. 265ms) reflects a trade-off for this improved pause handling. We note our evaluation is based on the agent text outputs and agent start and end timestamps are based on explicit agent \texttt{<bos>} and \texttt{<eos>} tokens in modeling.

We have also compared the proposed model to a baseline model without streaming ASR head for turn taking (Table \ref{tab:duplex_tt}), Intelligence (Table \ref{tab:duplex_intel}) and FDB-v1 (Table \ref{tab:fdb}), respectively. Overall, adding streaming ASR head does not signifcantly change the turn taking results compared to the baseline model (i.e., no ASR) as shown Table \ref{tab:duplex_tt} and \ref{tab:fdb}, except increasing the latency of the smooth turn taking set in FDB-v1. However, in a multi-turn conversation the turn taking latency remains similar (Table \ref{tab:duplex_tt}). On the other hand, adding the ASR head leads to the improvement in OpenbookQA (Table \ref{tab:duplex_intel}) from 66.59\% to 69.01\%, which indicates that the model may benefit from the text modality to answer questions.

\subsection{Standalone Streaming ASR}
\label{sect:standalone_asr}

We also train a standalone streaming ASR model using the same architecture but without the agent text heads, focusing solely on the streaming speech recognition task. Table~\ref{tab:asr_results} compares our standalone model against state-of-the-art streaming ASR systems on the HuggingFace Open ASR Leaderboard \cite{hf_open_asr_leaderboard}.

Our base model with 1.6s streaming delay achieves 8.47\% average WER on the HuggingFace Open ASR Leaderboard, and adding YODAS and YTC data from Granary \cite{koluguri2025granary} improves this to 7.73\%. Starting from this model, we have also done ablations regarding the streaming latency and achieved 7.99 \% average WER with a streaming latency of 1.2s. Regarding the LLM backbone, we have also tried a smaller backbone: Qwen 2.5-1.5B-Instruct \cite{qwen2025qwen25}, and achieved an average WER of 8.64\%.

We note that there are still gaps comparing our model to the SOTA streaming ASR models. For example, the remaining gap compared to Nemotron-Speech-0.6B (7.16\% vs 7.73\%) is likely due to utilizing subsets of the Granary dataset. We note that, at the time of training, some portions of the Granary data were not available in our training pipeline, and we plan to incorporate the full dataset in future work. Compared to other SOTA models such as Qwen3-ASR and Kyutai STT (Table \ref{tab:asr_results}), our model achieves a lower streaming latency, though at the cost of higher WER. A direct comparison is also difficult as the training data of these models differs from ours and is not fully disclosed.

\section{Conclusions}

We presented an efficient method to add streaming ASR capabilities to a full-duplex speech-to-speech model. By introducing a lightweight ASR head in parallel to the agent text head, our approach enables real-time user transcription without significantly modifying the base S2S architecture. The duplex S2S model with integrated ASR achieves 10.21\% average WER while maintaining competitive turn-taking, and barge-in performance. This enables applications such as conversation logging and accessibility features. Furthermore, we showed that the same architecture trained as a standalone streaming ASR model achieves 7.73\% WER on the HuggingFace Open ASR Leaderboard.



\section{Generative AI Use Disclosure}
Claude Opus 4.8 and Codex with GPT-5.5 are used to format tables and references and fix grammatical errors throughout all sections of the paper.

\bibliographystyle{IEEEtran}
\bibliography{ref}

@article{shih2025speech_llm_think,
  title={Can Speech {LLMs} Think while Listening?},
  author={Shih, Yi-Jen and Raj, Desh and Wu, Chunyang and Zhou, Wei and Bong, SK and Gaur, Yashesh and Mahadeokar, Jay and Kalinli, Ozlem and Seltzer, Mike},
  journal={arXiv preprint arXiv:2510.07497},
  year={2025}
}

@article{defossez2024moshi,
  title={Moshi: a speech-text foundation model for real-time dialogue},
  author={D{\'e}fossez, Alexandre and Mazar{\'e}, Laurent and Orsini, Manu and Royer, Am{\'e}lie and P{\'e}rez, Patrick and J{\'e}gou, Herv{\'e} and Grave, Edouard and Zeghidour, Neil},
  journal={arXiv preprint arXiv:2410.00037},
  year={2024}
}

@inproceedings{roy2026personaplex,
  title={PersonaPlex: Voice and Role Control for Full Duplex Conversational Speech Models},
  author={Roy, Rajarshi and Raiman, Jonathan and Lee, Sang-gil and Ene, Teodor-Dumitru and Kirby, Robert and Kim, Sungwon and Kim, Jaehyeon and Catanzaro, Bryan},
  booktitle={IEEE International Conference on Acoustics, Speech and Signal Processing (ICASSP)},
  year={2026},
  organization={IEEE}
}

@misc{nvidia2025nemotron,
  author       = {NVIDIA},
  title        = {{Nemotron-Nano-9B-v2-Base: A 9B Parameter Language Model for Reasoning and Instruction Following}},
  year         = {2025},
  url          = {https://huggingface.co/nvidia/NVIDIA-Nemotron-Nano-9B-v2-Base},
  note         = {Hugging Face Model Hub}
}

@article{brown2020language,
  title={Language models are few-shot learners},
  author={Brown, Tom B and Mann, Benjamin and Ryder, Nick and Subbiah, Melanie and Kaplan, Jared and Dhariwal, Prafulla and Neelakantan, Arvind and Shyam, Pranav and Sastry, Girish and Askell, Amanda and others},
  journal={arXiv preprint arXiv:2005.14165},
  year={2020}
}

@article{muralidharan2024minitron,
  title={LLM Pruning and Distillation in Practice: The Minitron Approach},
  author={Muralidharan, Saurav and Sreenivas, Sharath Turuvekere and Joshi, Raviraj and Chochowski, Marcin and Patwary, Mostofa and Shoeybi, Mohammad and Catanzaro, Bryan and Kautz, Jan and Molchanov, Pavlo},
  journal={arXiv preprint arXiv:2407.14679},
  year={2024}
}

@article{nvidia2025nemotron_nano2,
  title={NVIDIA Nemotron Nano 2: An Accurate and Efficient Hybrid Mamba-Transformer Reasoning Model},
  author={Bakhtin, Anton and Casper, Stephen and Chochowski, Marcin and Du, Jiaqi and Feinberg, Vladimir and Ganguli, Deep and Joshi, Raviraj and Kautz, Jan and Korneev, Alex and Kosson, Aran and others},
  journal={arXiv preprint arXiv:2508.14444},
  year={2025}
}

@article{openai2024gpt4o,
  title={GPT-4o System Card},
  author={OpenAI},
  journal={arXiv preprint arXiv:2410.21276},
  year={2024}
}

@misc{openai2025realtime,
  author = {OpenAI},
  title = {Introducing gpt-realtime and Realtime API updates for production voice agents},
  year = {2025},
  url = {https://openai.com/index/introducing-gpt-realtime/},
  note = {Blog post}
}

@article{qwen2025qwen3,
  title={Qwen3 Technical Report},
  author={Qwen Team},
  journal={arXiv preprint arXiv:2505.09388},
  year={2025}
}

@misc{anthropic2024claude3,
  author = {Anthropic},
  title = {The Claude 3 Model Family: Opus, Sonnet, Haiku},
  year = {2024},
  url = {https://www-cdn.anthropic.com/de8ba9b01c9ab7cbabf5c33b80b7bbc618857627/Model_Card_Claude_3.pdf},
  note = {Model Card}
}

@article{team2025gemini25,
  title={Gemini 2.5: Pushing the Frontier with Advanced Reasoning, Multimodality, Long Context, and Next Generation Agentic Capabilities},
  author={Team, Gemini and Anil, Rohan and Borgeaud, Sebastian and Casas, Jean-Baptiste and Fiedel, Noah and Georgiou, Konstantinos and Gulati, Anmol and Gu, Shixiang Shane and Hu, Hexiang and Kalashnikov, Dmitry and others},
  journal={arXiv preprint arXiv:2507.06261},
  year={2025}
}

@article{qwen2025omni,
  title={Qwen3-Omni Technical Report},
  author={Qwen Team and Xu, Kai and Zhang, Zihan and Wang, Xinyu and Dong, Yichao and Chen, Zirui and Zhou, Junyang and Lin, Junyang and Yang, An and Chu, Yunfei and others},
  journal={arXiv preprint arXiv:2509.17765},
  year={2025}
}

@article{qwen2026asr,
  title={Qwen3-ASR Technical Report},
  author={Qwen Team},
  journal={arXiv preprint arXiv:2601.21337},
  year={2026}
}

@article{wang2024speechllm,
  title={Recent Advances in Speech Language Models: A Survey},
  author={Wang, Wenqian and Yan, Daxin and Li, Ziyang and Li, Shuai and Tian, Qingqing and Chen, Xie},
  journal={arXiv preprint arXiv:2410.03751},
  year={2024}
}

@article{intrinsicvoice2024,
  title={IntrinsicVoice: Empowering LLMs with Intrinsic Real-time Voice Interaction Abilities},
  author={Xin, Xiang and Wang, Zheng and Cheng, Qun and Chen, Xiao and Li, Zizheng and Jiang, Xin and Zhao, Hai and Feng, Yang},
  journal={arXiv preprint arXiv:2410.08035},
  year={2024}
}

@article{llamaomni2024,
  title={LLaMA-Omni: Seamless Speech Interaction with Large Language Models},
  author={Fang, Qingkai and Guo, Shoutao and Zhou, Yan and Ma, Zhengrui and Zhang, Shaolei and Feng, Yang},
  journal={arXiv preprint arXiv:2409.06666},
  year={2024}
}

@article{fireredchat2024,
  title={FireRedChat: A Pluggable, Full-Duplex Voice Interaction System with Cascaded and Semi-Cascaded Implementations},
  author={Chen, Yiwei and Hu, Taihao and Li, Yixuan and Tang, Yuping and Su, Hang and Zheng, Xiaoyu and Lin, Zhengyang and Wu, Sicheng and Zhang, Jun and Zhou, Joey Tianyi},
  journal={arXiv preprint arXiv:2509.06502},
  year={2024}
}

@article{flexduo2025,
  title={FlexDuo: A Pluggable System for Enabling Full-Duplex Capabilities in Speech Dialogue Systems},
  author={Zhang, Ziyang and Chen, Jiahao and Liu, Yuanfeng and Li, Hao and Zhang, Yifan and Lin, Ziqiang and Zhou, Shiyu and Zhang, Wei-Qiang and Liu, Jia},
  journal={arXiv preprint arXiv:2502.13472},
  year={2025}
}

@article{pratap2024scaling,
  title={Scaling speech technology to 1,000+ languages},
  author={Pratap, Vineel and Tjandra, Andros and Shi, Bowen and Tomasello, Paden and Babu, Arun and Kundu, Sayani and Elkahky, Ali and Ni, Zhaoheng and Vyas, Apoorv and Fazel-Zarandi, Maryam and others},
  journal={Journal of Machine Learning Research},
  volume={25},
  number={97},
  pages={1--52},
  year={2024}
}

@article{yusalmonn,
  title={SALMONN-omni: A Codec-free LLM for Full-duplex Speech Understanding and Generation},
  author={Yu, Wenyi and Wang, Siyin and Yang, Xiaoyu and Chen, Xianzhao and Tian, Xiaohai and Zhang, Jun and Sun, Guangzhi and Lu, Lu and Wang, Yuxuan and Zhang, Chao},
  journal={arXiv preprint arXiv:2411.18138},
  year={2024}
}

@article{dubey2024llama,
  title={The llama 3 herd of models},
  author={Dubey, Abhimanyu and Jauhri, Abhinav and Pandey, Abhinav and Kadian, Abhishek and Al-Dahle, Ahmad and Letman, Aiesha and Mathur, Akhil and Schelten, Alan and Yang, Amy and Fan, Angela and others},
  journal={arXiv preprint arXiv:2407.21783},
  year={2024}
}

@misc{nvidia2023stt_fastconformer,
  author       = {NVIDIA},
  title        = {{STT En FastConformer Hybrid Transducer-CTC Large Streaming 80ms}},
  year         = {2023},
  url          = {https://catalog.ngc.nvidia.com/orgs/nvidia/teams/nemo/models/stt_en_fastconformer_hybrid_large_streaming_80ms},
  note         = {Version 1.20.0, Released June 22, 2023}
}

@misc{nvidia2023stt_fastconformer_multi,
  author       = {NVIDIA},
  title        = {{STT En FastConformer Hybrid Transducer-CTC Large Streaming Multi}},
  year         = {2023},
  url          = {https://huggingface.co/nvidia/stt_en_fastconformer_hybrid_large_streaming_multi},
  note         = {Hugging Face Model Hub}
}

@article{zhang2024omniflatten,
  title={Omniflatten: An end-to-end gpt model for seamless voice conversation},
  author={Zhang, Qinglin and Cheng, Luyao and Deng, Chong and Chen, Qian and Wang, Wen and Zheng, Siqi and Liu, Jiaqing and Yu, Hai and Tan, Chaohong and Du, Zhihao and others},
  journal={arXiv preprint arXiv:2410.17799},
  year={2024}
}

@article{wang2024freeze,
  title={Freeze-omni: A smart and low latency speech-to-speech dialogue model with frozen llm},
  author={Wang, Xiong and Li, Yangze and Fu, Chaoyou and Shen, Yunhang and Xie, Lei and Li, Ke and Sun, Xing and Ma, Long},
  journal={arXiv preprint arXiv:2411.00774},
  year={2024}
}

@article{xie2024mini,
  title={Mini-omni2: Towards open-source gpt-4o with vision, speech and duplex capabilities},
  author={Xie, Zhifei and Wu, Changqiao},
  journal={arXiv preprint arXiv:2410.11190},
  year={2024}
}

@article{guo2025deepseek,
  title={Deepseek-r1: Incentivizing reasoning capability in llms via reinforcement learning},
  author={Guo, Daya and Yang, Dejian and Zhang, Haowei and Song, Junxiao and Zhang, Ruoyu and Xu, Runxin and Zhu, Qihao and Ma, Shirong and Wang, Peiyi and Bi, Xiao and others},
  journal={arXiv preprint arXiv:2501.12948},
  year={2025}
}

@article{deepseek2025v32,
  title={DeepSeek-V3.2: Pushing the Frontier of Open Large Language Models},
  author={DeepSeek-AI and Guo, Daya and Qin, Dejian and Fan, Zhenda and Liu, Zhibin and Ruan, Xuyang and Liang, Wangding and Shi, Yuxiang and Guo, Qihao and Shao, Zhenda and others},
  journal={arXiv preprint arXiv:2512.02556},
  year={2025}
}

@article{minimax2025m01,
  title={MiniMax-01: Scaling Foundation Models with Lightning Attention},
  author={MiniMax and Deng, Shiyu and Yao, Qiyao and Jia, Jianxin and Zhang, Xiaokang and Wang, Jingwei and Wu, Haoyu and Han, Xu and Zhang, Yue and Xu, Wei and others},
  journal={arXiv preprint arXiv:2501.08313},
  year={2025}
}

@inproceedings{huang2024audiogpt,
  title={Audiogpt: Understanding and generating speech, music, sound, and talking head},
  author={Huang, Rongjie and Li, Mingze and Yang, Dongchao and Shi, Jiatong and Chang, Xuankai and Ye, Zhenhui and Wu, Yuning and Hong, Zhiqing and Huang, Jiawei and Liu, Jinglin and others},
  booktitle={Proceedings of the AAAI Conference on Artificial Intelligence},
  volume={38},
  number={21},
  pages={23802--23804},
  year={2024}
}

@article{zhang2023speechgpt,
  title={Speechgpt: Empowering large language models with intrinsic cross-modal conversational abilities},
  author={Zhang, Dong and Li, Shimin and Zhang, Xin and Zhan, Jun and Wang, Pengyu and Zhou, Yaqian and Qiu, Xipeng},
  journal={arXiv preprint arXiv:2305.11000},
  year={2023}
}

@article{kim2024unified,
  title={Unified Speech-Text Pretraining for Spoken Dialog Modeling},
  author={Kim, Heeseung and Seo, Soonshin and Jeong, Kyeongseok and Kwon, Ohsung and Kim, Jungwhan and Lee, Jaehong and Song, Eunwoo and Oh, Myungwoo and Yoon, Sungroh and Yoo, Kang Min},
  journal={arXiv preprint arXiv:2402.05706},
  year={2024}
}

@article{zeng2024glm,
  title={Glm-4-voice: Towards intelligent and human-like end-to-end spoken chatbot},
  author={Zeng, Aohan and Du, Zhengxiao and Liu, Mingdao and Wang, Kedong and Jiang, Shengmin and Zhao, Lei and Dong, Yuxiao and Tang, Jie},
  journal={arXiv preprint arXiv:2412.02612},
  year={2024}
}

@inproceedings{chen2024salm,
  title={Salm: Speech-augmented language model with in-context learning for speech recognition and translation},
  author={Chen, Zhehuai and Huang, He and Andrusenko, Andrei and others},
  booktitle={ICASSP},
  pages={13521--13525},
  year={2024},
  organization={IEEE}
}

@article{parakeet,
  title={Parakeet: A Natural Language Speech Recognition Model},
  author={Sridhar, Sravya and Puvvada, Krishna C and Chen, Zhehuai and Hrinchuk, Oleksii and Huang, He and Lavrukhin, Vitaly and Balam, Jagadeesh and Ginsburg, Boris},
  journal={NVIDIA Technical Blog},
  year={2024},
  url={https://nvidia.github.io/NeMo/blogs/2024/2024-01-parakeet/}
}

@article{qwenasr,
  title={Qwen2-Audio Technical Report},
  author={Chu, Yunfei and Xu, Jin and Yang, Qian and Wei, Haojie and Wei, Xipin and Guo, Zhifang and Leng, Yichong and Lv, Yuanjun and He, Jinzheng and Lin, Junyang and Zhou, Chang and Zhou, Jingren},
  journal={arXiv preprint arXiv:2407.10759},
  year={2024}
}

@article{salm_duplex,
  title={SALM-Duplex: Efficient and Direct Duplex Modeling for Speech-to-Speech Language Model},
  author={Hu, Ke and Hosseini-Asl, Ehsan and Chen, Chen and Casanova, Edresson and Ghosh, Subhankar and {\.Z}elasko, Piotr and Chen, Zhehuai and Li, Jason and Balam, Jagadeesh and Ginsburg, Boris},
  journal={arXiv preprint arXiv:2505.15670},
  year={2025}
}

@inproceedings{asru_duplex,
  title={Open Full-duplex Voice Agent with Speech-to-Speech Language Model},
  author={Casanova, Edresson and Chen, Chen and Hu, Kevin and Pasad, Ankita and Rastorgueva, Elena and Narasimhan, Seelan Lakshmi and Deng, Slyne and Hosseini-Asl, Ehsan and {\.Z}elasko, Piotr and Mendelev, Valentin and Ghosh, Subhankar and Peng, Yifan and Chen, Zhehuai and Li, Jason and Balam, Jagadeesh and Lavrukhin, Vitaly and Ginsburg, Boris},
  booktitle={ASRU},
  year={2025}
}

@article{voicebench,
  title={VoiceBench: Benchmarking LLM-Based Voice Assistants},
  author={Chen, Yiming and Yue, Xiangyu and Zhang, Chen and Gao, Xiaoxue and Tan, Robby T and Li, Haizhou},
  journal={arXiv preprint arXiv:2410.17196},
  year={2024}
}

@inproceedings{panayotov2015librispeech,
  title={Librispeech: an ASR corpus based on public domain audio books},
  author={Panayotov, Vassil and Chen, Guoguo and Povey, Daniel and Khudanpur, Sanjeev},
  booktitle={2015 IEEE International Conference on Acoustics, Speech and Signal Processing (ICASSP)},
  pages={5206--5210},
  year={2015},
  organization={IEEE}
}

@article{koluguri2025granary,
  title={Granary: Speech Recognition and Translation Dataset in 25 European Languages},
  author={Koluguri, Nithin Rao and Sekoyan, Monica and Zelenfroynd, George and Meister, Sasha and Ding, Shuoyang and Kostandian, Sofia and Huang, He and Karpov, Nikolay and Balam, Jagadeesh and Lavrukhin, Vitaly and Peng, Yifan and Papi, Sara and Gaido, Marco and Brutti, Alessio and Ginsburg, Boris},
  journal={arXiv preprint arXiv:2505.13404},
  year={2025}
}

@misc{hf_open_asr_leaderboard,
  author = {Hugging Face},
  title = {Open ASR Leaderboard},
  year = {2024},
  url = {https://huggingface.co/spaces/hf-audio/open_asr_leaderboard},
  note = {Hugging Face Space}
}

@inproceedings{nvidia2025nemotron_speech_streaming,
  title={Stateful Conformer with Cache-based Inference for Streaming Automatic Speech Recognition},
  author={Noroozi, Vahid and Majumdar, Somshubra and Kumar, Ankur and Balam, Jagadeesh and Ginsburg, Boris},
  booktitle={ICASSP},
  year={2024},
  organization={IEEE}
}

@misc{kyutai2024stt,
  author       = {Kyutai},
  title        = {{STT-2.6b-en: Streaming Speech-to-Text Model}},
  year         = {2024},
  url          = {https://huggingface.co/kyutai/stt-2.6b-en},
  note         = {Hugging Face Model Hub}
}

@article{qwen2025qwen25,
  title={Qwen2.5 Technical Report},
  author={Qwen Team},
  journal={arXiv preprint arXiv:2412.15115},
  year={2025}
}

@inproceedings{wang2021voxpopuli,
  title={{VoxPopuli}: A Large-Scale Multilingual Speech Corpus for Representation Learning, Semi-Supervised Learning and Interpretation},
  author={Wang, Changhan and Riviere, Morgane and Lee, Ann and Wu, Anne and Talnikar, Chaitanya and Haziza, Daniel and Schwab, Mary and Pino, Juan and Dupoux, Emmanuel},
  booktitle={Proceedings of the 59th Annual Meeting of the Association for Computational Linguistics},
  pages={993--1003},
  year={2021}
}

@inproceedings{ardila2020common,
  title={Common Voice: A Massively-Multilingual Speech Corpus},
  author={Ardila, Rosana and Branson, Megan and Davis, Kelly and Kohler, Michael and Meyer, Josh and Henretty, Michael and Morais, Reuben and Saunders, Lindsay and Tyers, Francis and Weber, Gregor},
  booktitle={Proceedings of the 12th Language Resources and Evaluation Conference},
  pages={4218--4222},
  year={2020}
}

@inproceedings{yamagishi2019vctk,
  title={{CSTR VCTK} Corpus: English Multi-speaker Corpus for {CSTR} Voice Cloning Toolkit},
  author={Yamagishi, Junichi and Veaux, Christophe and MacDonald, Kirsten},
  booktitle={University of Edinburgh. The Centre for Speech Technology Research},
  year={2019}
}

@article{oneill2021spgispeech,
  title={{SPGISpeech}: 5,000 Hours of Transcribed Financial Audio for Fully Formatted End-to-End Speech Recognition},
  author={O'Neill, Patrick K and Lavrukhin, Vitaly and Majumdar, Somshubra and Noroozi, Vahid and Zhang, Yuekai and Kuchaiev, Oleksii and Balam, Jagadeesh and Huang, Yuliya and Krivoshein, Aleksandr and Ginsburg, Boris},
  journal={arXiv preprint arXiv:2104.02014},
  year={2021}
}

@article{casanova2026voicechat,
  title={VoiceChat-TTS: A Low-Latency Continuous Speech Synthesis Model for Interactive Agents},
  author={Casanova, Edresson and Kim, Jaehyeon and Fuenmayor, Mariana Graterol and Hussain, Shehzeen and Klimkov, Viacheslav and Mendelev, Valentin and Desta, Mikyas and Neekhara, Paarth and Zelasko, Piotr and Chen, Chen and others},
  journal={arXiv preprint arXiv:2608.13831},
  year={2026}
}

@article{lin2025fullduplexbench,
  title={Full-Duplex-Bench: A Benchmark to Evaluate Full-duplex Spoken Dialogue Models on Turn-taking Capabilities},
  author={Lin, Guan-Ting and Lian, Jiachen and Li, Tingle and Wang, Qirui and Anumanchipalli, Gopala and Liu, Alexander H and Lee, Hung-yi},
  journal={arXiv preprint arXiv:2503.04721},
  year={2025}
}

@misc{silero_vad,
  author       = {{Silero Team}},
  title        = {Silero {VAD}: pre-trained enterprise-grade Voice Activity Detector},
  year         = {2021},
  howpublished = {\url{https://github.com/snakers4/silero-vad}},
  note         = {GitHub repository}
}

\end{document}